\documentclass{article}
\usepackage{graphicx} 
\usepackage{todonotes}
\usepackage{caption}
\usepackage{subcaption}
\usepackage{url}
\usepackage{hyperref}

\title{Yin Yang Convolutional Nets: Image Manifold
Extraction by the Analysis of Opposites}
\author{
Augusto Seben da Rosa\textsuperscript{1},
Frederico Santos de Oliveira\textsuperscript{2}, \\
Anderson da Silva Soares\textsuperscript{3}, 
Arnaldo Candido Junior\textsuperscript{4}
 \\ \\
\textsuperscript{1} Federal University of Technology - Paraná \\
\textsuperscript{2} Federal University of Mato Grosso \\
\textsuperscript{3} Federal University of Goias \\
\textsuperscript{4} São Paulo State University \\
}
\date{}

\begin{document}

\maketitle

\begin{abstract}
Computer vision in general presented several advances such as training optimizations, new  architectures (pure attention, efficient block, vision language models, generative models, among others). This have improved performance in several tasks such as classification, and others. However, the majority of these models focus on modifications that are taking distance from realistic neuroscientific approaches related to the brain. In this work, we adopt a more bio-inspired approach and present the Yin Yang Convolutional Network, an architecture that extracts visual manifold, its blocks are intended to separate analysis of colors and forms at its initial layers, simulating occipital lobe's operations. Our results shows that our architecture provides State-of-the-Art efficiency among low parameter architectures in the dataset CIFAR-10. Our first model reached 93.32\% test accuracy, 0.8\% more than the older SOTA in this category, while having 150k less parameters (726k in total). Our second model uses 52k parameters, losing only 3.86\% test accuracy. We also performed an analysis on ImageNet, where we reached 66.49\% validation accuracy with 1.6M parameters. We make the code publicly available at: \href{https://github.com/NoSavedDATA/YinYang_CNN}{\url{https://github.com/NoSavedDATA/YinYang_CNN}}.
\end{abstract}

\section{Introduction}

The field of neural computer vision have presented a great advancement,  for instance, improved neural network architectures were proposed. Some of these architectures include: mobile model families as MobileNet \cite{mobilenetv3}, EfficientNet V2 \cite{efficientnetv2} and RegNet \cite{regnet}; two-branch neural networks for semantic segmentation, as BiSeNet \cite{bisenet}, Deep Dual-Resolution networks \cite{deepdual} and SeaFormer \cite{seaformer}; pure attention mechanisms applied into image classification, such as ViT \cite{vit} and MaxViT \cite{maxvit}; image generative models like Stable Diffusion \cite{stablediffusion} and DALL-E-2 \cite{dalle2}; and lastly, vision-language models, as CLIP \cite{clip}. The majority of these architectures focuses on increasing model efficiency by improving  the micro-architecture -- there is, by making adjustments relative to the inside of a network block, as in mobile model families. Some of these architecture also leverage the potential of CNNs and Transformers into high level computer vision tasks, as image generation or vision-language models creation. 

However, none of the famous modern architectures are trying to reach a more realistic neuroscientific approach in respect to brain occipital (visual cortex) mechanisms. In this regard, we base our research on two neuroscientific findings about function specialized occipital lobe areas, that is, edge detection that happens on V1 area \cite{V1_Edge} and color processing at V4 area \cite{V4_Color}. Also, this form of  specialization is also observed in the human eye, in which rod components are related to white and black processing \cite{eye_rod}, and cone components to color processing \cite{eye_cone}.

In this research, we present Yin Yang Convolutional Net (YYNet), a neural network model which makes adjustments relative to the global scale of the model. This is performed in the macro-architecture by aggregating blocks (or single purpose networks) that extracts visual manifolds by doing a separate analysis of colors and forms from its input. We find that this architecture provides State-of-the-Art (SOTA) efficiency among low parameter architectures applied to the low data and image resolution dataset, CIFAR-10, using less parameters and less training epochs than existing models.

\section{Related Work}

Vision neural network research have grown in quality at a very fast pace. In this section, we present related architectures and their given tasks. Related work can be categorized into five approaches: mobile networks; two branch networks; transformers; vision-language models; generative models.

Some of this approaches are usually divided in three parts: stem, stage and head. The stem is usually single convolution with stride 2, but may present optional extra convolutions. The stage contains the main architecture of the model, that can be divided into blocks with layers, in which each stage block shares hyperparameters (number of channels and extra hyperparameters) across all its layers. The head may or not contain convolutions and then it is followed by average pooling, an optional linear layer and the final classification linear layer followed by a softmax. This is the case for Mobile Nets and MaxViT.

\subsection{Mobile Networks}

\begin{figure}
    \centering
    \begin{subfigure}{0.55\textwidth}
         \centering
         \includegraphics[width=\textwidth]{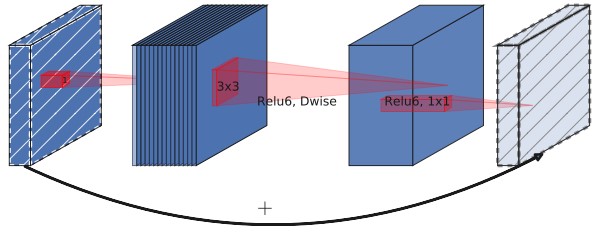}
         \caption{Mobile Net V2}
         \label{fig:mbnet2}
    \end{subfigure}
    \begin{subfigure}{0.55\textwidth}
         \centering
         \includegraphics[width=\textwidth]{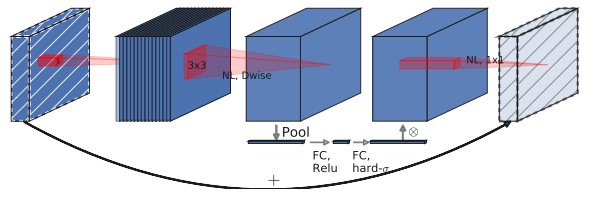}
         \caption{Mobile Net V3}
         \label{fig:mbnet3}
    \end{subfigure}
    \caption{Mobile Net Micro-Architectures \cite{mobilenetv3}}
    \label{fig:MobileNets}
\end{figure}

Mobile network based approaches focus on building efficient models with fewer parameters. The authors of MobileNet V2 \cite{mobilenetv2}, shown in Figure \ref{fig:MobileNets}, propose residual inverted bottleneck blocks, a micro-architecture that became more efficient than the reference model ResNet \cite{resnet}. The objective of this micro-architecture is to reduce data dimensionality in a way that the manifold spans the entire space of lower dimensional sub-spaces. They do so by inserting inverted bottlenecks at each block, instead of keeping the channels number constant at repeating blocks as in ResNet. That is, they first expand the number of channels for a given factor (e.g.  4), then they apply convolutions in this higher dimension and finally go back into the original dimension, similar to feed-forward networks in the transformer architecture \cite{vaswani2017attention}. On their architecture, they first increase the channels number with 1x1 kernels, then use 3x3 depth-wise kernels on the higher dimension and, lastly, they reduce the dimension back to what it was before with a 1x1 kernel. The reason to use 1x1 kernels follwed by 3x3 is because the authors of Mobile Net V1 \cite{mobilenet} found it more efficient
than directly  expanding using 3x3 kernels.

Further, on Mobile Net V3 \cite{mobilenetv3}, they improve Mobile Net V2 efficiency by applying a Squeeze and Excitation \cite{squeeze_excitation} mechanism after the 3x3 depth-wise convolution. We will refer to the block of Mobile Net V3 as the MBConv.

Besides, in EfficientNet V2 \cite{efficientnetv2}, the authors have changed the original architecture of MobileNet V3 into the Fused MBConv and interpolated this new block with the original MBConv, finding the best parameter configuration given their applied model search. They have also proposed slight modifications to the compound scaling method of EfficientNet \cite{efficientnet}, which consists in adjusting model depth, channel number and image size to find the best scale inside a family of models.

\subsection{Two-branch Neural Networks}
\label{sec:vision}

\begin{figure}[ht]
\centering
\includegraphics[width=0.4\textwidth]{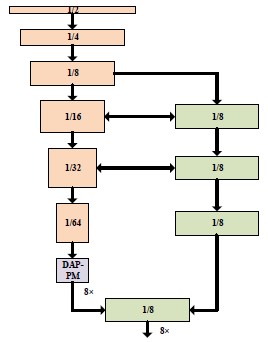}
\caption{Deep Dual-Resolution Network \cite{deepdual}.}
\label{fig:deep_dual}
\end{figure}

Two-branch based models allow the signal to travel along two different paths, conventionally after a shared stem - except for BiSeNet. This is done to extract manifold efficiently. An example of different paths for signal propagation can be seem  in BiSeNet \cite{bisenet} and Deep Dual-Resolution Networks \cite{deepdual}, these architectures are built for the purpose of semantic segmentation, in which one of the branches is shallow and wide, extracting local details, and the other is deep and narrow, capturing high-level semantics. For simplicity, we show only the Deep Dual-Resolution Network at Figure \ref{fig:deep_dual}.

\textbf{}

\begin{figure}[ht]
\centering
\includegraphics[width=0.99\textwidth]{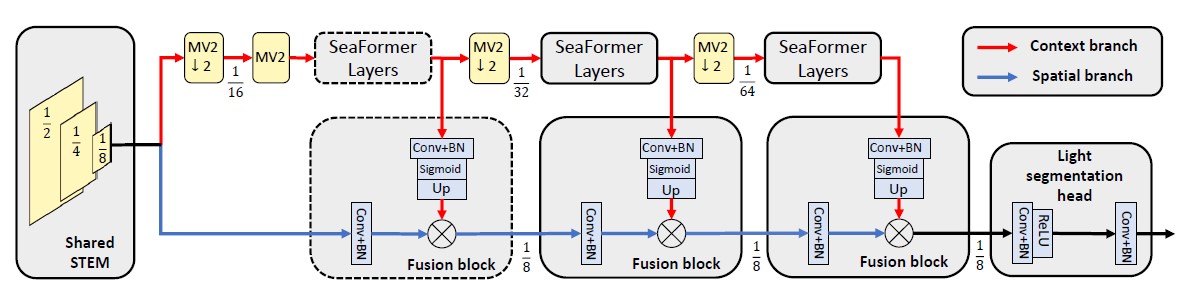}
\caption{SeaFormer \cite{seaformer}.}
\label{fig:sea_former}
\end{figure}

Another example of two branch network is SeaFormer \cite{seaformer}, that increases the efficiency by using mobile transformers, from which the embedding of the stem backbone is processed and fused at multiple steps with mobile transformer blocks embeddings, as can be seen at Figure \ref{fig:sea_former}. Our work is inspired by these architectures.

\subsection{Vision Transformers}

\begin{figure}[ht]
\centering
\includegraphics[width=0.70\textwidth]{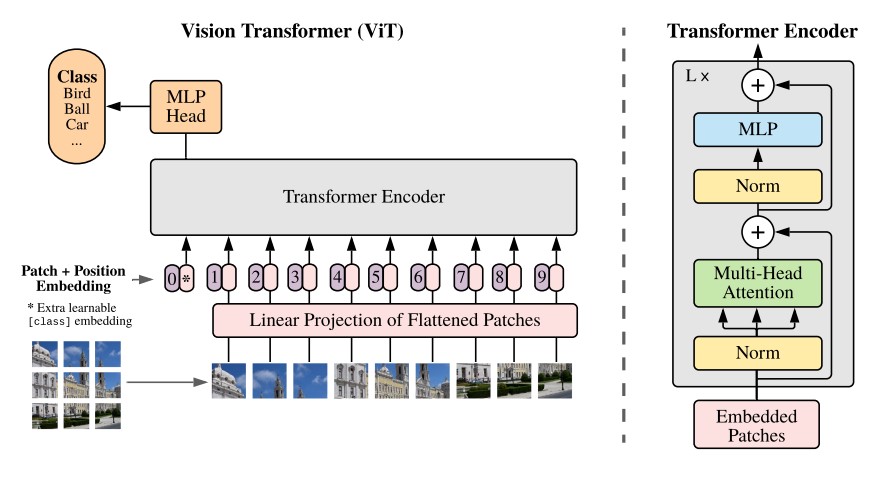}
\caption{ViT \cite{vit}.}
\label{fig:vit}
\end{figure}

Vision Transformers apply transformer encoder blocks to explore the attention paradigm into vision tasks. On ViT \cite{vit}, as shown in Figure \ref{fig:vit}, the authors do this by dividing the image into multiple embedding patches of equal window size, in which each patch embedding is cross-attended with attention to all other patches. They also reserve a patch embedding for classification (usually referred as ``cls'' token in language models), which does not come from the image, and so aggregates global information about other patches. The advantage of using a pure transformers architecture is that the multiple heads mechanism turns it possible to apply tensor parallelism \cite{megatron_llm}.

Then, MaxViT \cite{maxvit} enhances ViT with more efficient attention mechanisms for the vision paradigm, the authors use Axial Attention for local details and Grid Attention for global interactions between pixels. They also append a MBConv at the start of each block.

%\subsection{Other Approaches}

%Two other approaches for computer vision problems are Vision-Language models and Generative models. Vision-Language Models (VLM) are macro-architectures able to produce multimodal embeddings or manifolds. The authors of CLIP \cite{clip} train 2 networks, one vision encoder, ResNet or ViT, and a transformer text encoder. The final embedding of each network composes the loss in a contrastive learning method. The objective is to maximize the embeddings cosine similarity of correctly matched pairs from both modalities on the batch and to minimize the cosine similarity of unmatched pairs.

%Generative Models were one of the Artificial Intelligence branches that have most impacted society on 2022. They generate images from pure noise image and text prompt inputs. At DALL-E-2 \cite{dalle2}, the authors use a CLIP text embedding to feed an autoregressive or diffusion prior to produce an image embedding, which is then used to condition a diffusion model. Stable Diffusion uses a U-Net \cite{unet} conditioned on transformer text embeddings prompt.

\section{Architecture}

\begin{figure}
    \centering
    \caption{Micro-architecture}
    \label{fig:mic}
    \begin{subfigure}[b]{0.2\textwidth}
        \includegraphics[width=\textwidth]{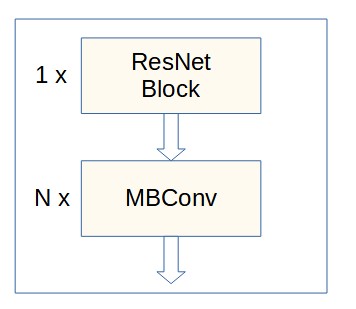}
        \caption{Micro-arch}
        \label{fig:mic_simp}
    \end{subfigure}
    \begin{subfigure}[b]{0.19\textwidth}
         \includegraphics[width=\textwidth]{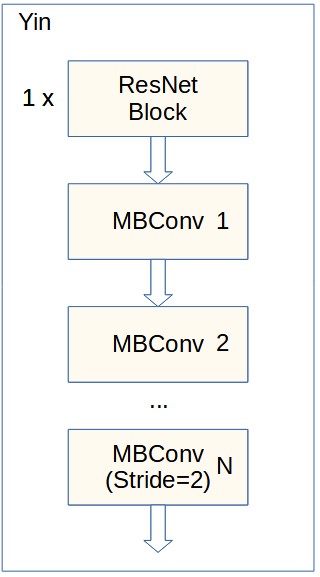}
         \caption{Yin}
         \label{fig:mic_yin}
    \end{subfigure}
    \begin{subfigure}[b]{0.19\textwidth}
         \includegraphics[width=\textwidth]{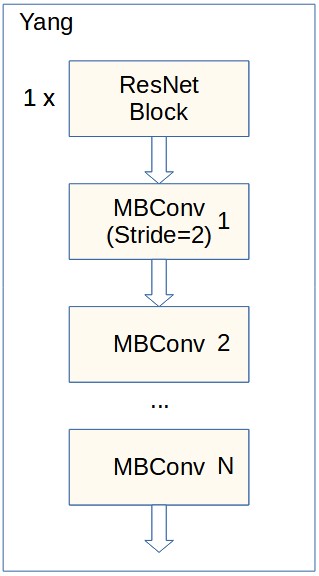}
         \caption{Yang}
         \label{fig:mic_yang}
    \end{subfigure}
    %\quad
    \begin{subfigure}[b]{0.194\textwidth}
         \includegraphics[width=\textwidth]{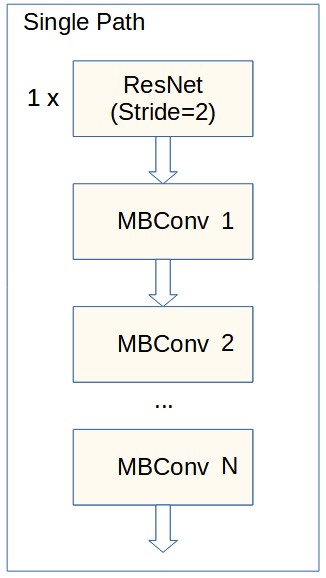}
         \caption{Single Path}
         \label{fig:mic_sp}
    \end{subfigure}
\end{figure}

Yin-Yang Net uses a \textbf{micro-architcture} presented in Figure \ref{fig:mic}. In each repeating block, we start with a sub-block of ResNet and then use $n$ sub-blocks of MBConv from Mobile Net V3. We found this configuration has better accuracy when training solely with MBConv given our hyperparameters, and it is more  efficient during training than using only ResNet blocks. At the two branch layers, stride 2 is applied on the last or the first sub-block according to the micro-architecture type, Yin or Yang, respectively. The single branch layers use stride 2 on the ResNet block.

% pacote subfigure

Our work is inspired by two branch architectures. However, our approach differs from classical two branch networks in 2 aspects. First, instead of building a shallow and a deep branch for details and semantics extraction, YYNet uses the same number of layers and channels at blocks on the same level, but stride 2 at different parts of these layers. 
%- this allows us to use the same amount of channels and layers on both branches for an easier hyperparametrization. 
Second, in our work, there is no common stem backbone to the branches, the stem is the focus of this paper, where different manifolds are analyzed. %(III) By not being a semantic segmentation network, but a classifier network, the main idea is to create a network to extract the most compact manifold with the most representative power; so, the focus is on the reducing of data dimensionality.

The Yin branch has the purpose of form analysis. Yin blocks can use the first channel of the input or the mean of all channels. We found that both configurations perform well. For simplicity, when the first channel approach is in use, the first block receives the red channel in the network input layer. This way, as there is no other color to extract, it is obliged to the task of extracting the form manifold. Also, in order to focus on local/higher scale details, a strategy of later stride 2 is used, meaning that the last MBConv of each block applies striding.

On the other hand, the Yang branch analyzes colors. With that in mind, as colors in nearly pixel are generally the same, we use an early stride 2 on this branch to remove color redundancy and only care about different colors interactions. That is, the first MBConv on each block applies striding. This block resembles standard single branch architectures, as its input is the three RGB channels and early stride 2 is applied.

\begin{figure}
\centering
\includegraphics[width=0.42\textwidth]{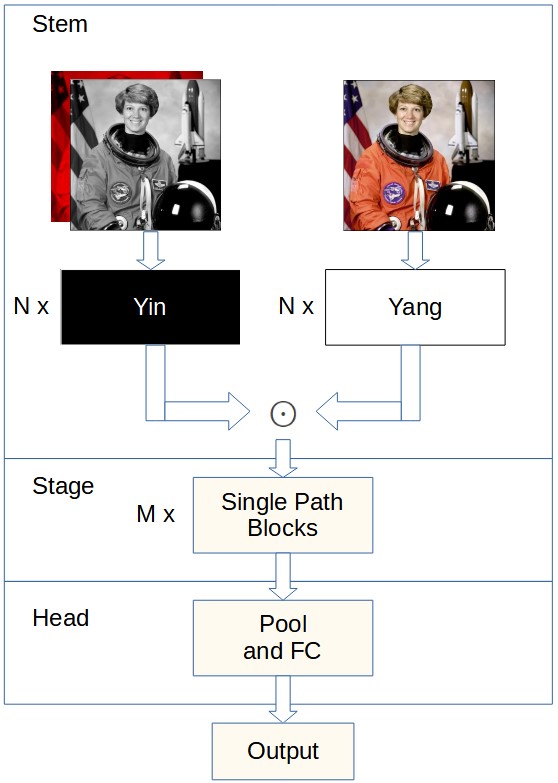}
\caption{Macro-architecture.}
\label{fig:mac}
\end{figure}

Then, at the \textbf{macro-architecture} level, as shown in Figure \ref{fig:mac}, we send the same input to both these micro-architectures. Their final embeddings have the same shape, as they have the same number  of layers and channels, with the exception of the first sub-block of each micro-architecture. We then apply a Fusion Gate mechanism adapted for our embeddings.

We apply a Fusion Gate, similar to SeaFormer and Multimodal Chain-of-Thought \cite{multimodalcot}. For this, we use an embedding fusion mechanism at outputs from Yin and Yang branches, each branch having a different embedding meaning. This is done to unify both embeddings, or the manifold, as presented in (\ref{eq:yin_yang}).

\begin{equation}
SP_X = A_Y + I_Y 
\label{eq:yin_yang}
\end{equation}

% multiplicação é um realce das cores internas
% subtração é outro realce de cores internas
% combinamos os dois para ter um realce ainda mais forte de cores internas

% porque fortalecer cores internas ajuda no processo de classificação?

$X$ represents the input of one network and $Y$ represents an output, $A$ represents the Yang blocks, $I$ represents the Yin blocks, \( \odot \) represents a Hadamard product (elementwise product). We have tested several combinations of $A$ and $I$ (as described in Section \ref{sec:exp}). The operations presented in (\ref{eq:yin_yang}) are those with best performance.
%On the first part: \(A_Y*(1-I_Y)\), we pretend to revert the outputs of the Yin network. This way, taking note that this network analysis forms and shapes, we hypothesize that the manifold matrix position where it outputs a number close to 1 is where it has identified an image outline. For that same position, there will be a color on the manifold of the Yang branch. Since we don't want to analyze colors on the image outlines for the proceeding manifold (the single branch network input), we reverse the Yin output by subtracting it by 1, thus guaranteeing the single branch input's manifold will contain an analysis of simplified color groups; or colors of whole regions. Then, on the second part of the formula: the residual connections, we apply a standard residual connection for the Yang output and reverse the Yin output accordingly to the first part of the formula. 
On CIFAR-10, this gated fusion yielded a slightly better accuracy than the concatenation – about less than 0.5\%, yet halving the channels number compared to concatenation.
%, with 40 epochs
%However, there is still need of a model search and hyperparameter tuning at ImageNet 1k.

After this, we send the embedding into Single Path blocks, that consists of a sub-block of ResNet with stride 2 on the first convolution and then $n$ sub-blocks of MBConv with no stride 2 (except for CIFAR-10, as described in Section \ref{sec:exp}). We use GELU \cite{gelu} as the activation function of any sub-block. As the head of our model, we use average pooling, flattening, a linear layer, GELU, dropout and the final classification linear layer followed by a softmax.

\section{Experiments and Results}
\label{sec:exp}

\subsection{Experiments}
We tested multiple fusion approaches to combine Yin and Yang outputs. We selected the approach presented in Equation \ref{eq:yin_yang}. However, Table \ref{tab:formulas} presents other approaches tested in this work. We performed 3 runs with batch size 512 on CIFAR-10 for each approach and report the mean and standard deviation.

\begin{table}[htb]
\centering
\caption{Fusion Approach}
\label{tab:formulas}
\begin{tabular}{l r r} 
 \hline
 \textbf{Formula} & \textbf{Mean} & \textbf{STD} \\ [0.5ex] 
 \hline
 A*(1-I) & 87.61 & 0.09  \\ 
 A*I + A+I & 87.63 & 0.08 \\
 A*(1-I) + A-I & 87.81 & 0.19 \\
 A*I & 87.87 & 0.23 \\
 A*(1-I) + A+I & 87.98 & 0.22 \\
 A+I & 88.21 & 0.46 \\ 
 \hline
\end{tabular}
\end{table}

We then reduced the batch size into 64 for the final model on CIFAR-10. We do this because batch normalization seems to work better in batch sizes on the range of 50 to 100 \cite{zhang2023dive}. We use a smaller batch size in ImageNet due to computational constraints.

We used gradient clip of 1 and mixed precision. At 25\% of training, we activate a exponential moving average with a multiplier of 0.1 for the averaged model parameter and 0.9 for current model parameter. Also, one of the graphs at \cite{adamw} shows that there are specific values of weight decay that works better with specific values of learning rate. We therefore use an adaptive value for the weight decay. At the end of each epoch, we set the weight decay to be equal to the learning rate*1.56. The input resolution for CIFAR-10 is 32x32 and ImageNet is 224x224. Further hyperparameters and settings are provided in Table \ref{tab:hyperparams}. Specific Hyperparameters adjustments for CIFAR-10 and ImageNet are presented on Table \ref{tab:arch_hyperparams}. We designed 3 models for CIFAR-10 and one for ImageNet. 
%There is a main difference between CIFAR-10 and ImageNet models.
%We reach similar accuracy over the related work focusing on small models.

\begin{table}[htb]
\centering
\caption{General Hyperparameters and Settings}
\label{tab:hyperparams}
\begin{tabular}{l r r} 
 \hline
 \textbf{Hyperparameter} & \textbf{YYNet Small} & \textbf{YYNet} \\ [0.5ex] 
 \hline
 Optimizer & AdamW \cite{adamw} & AdamW  \\ 
 LR Scheduler & One Cycle & One Cycle \\
 Max LR & 1e-2 & 18e-4  \\ 
 Epochs & 40 & 300  \\ 
 Batch Size & 64 & 32  \\ 
 GPU & RTX 2060 & RTX 2080 TI  \\ 
 Dataset & CIFAR-10 & ImageNet  \\
 \hline
\end{tabular}
\end{table}

\begin{table}[htb]
\centering
\caption{Dataset Specific Hyperparameters}
\label{tab:arch_hyperparams}
\begin{tabular}{l r r} 
 \hline
 \textbf{Hyperparameter} & \textbf{CIFAR-10} & \textbf{ImageNet} \\ [0.5ex] 
 \hline
 YY Starting Channels & (16, 32, 64) & 16  \\ 
 SP Starting Channels & (16, 32, 64) & 64 \\
 Channels added per MBConv & 0 & 2  \\
 Extra SP stride 2 & Yes & No  \\
 YY Layers & 1 & 1  \\
 SP Layers & 1 & 4  \\
 YY MBConv per Layer & 3 & 3  \\
 SP MBConv per Layer & 2 & 2  \\
 Pre-Classification Linear Neurons & 40 & 500  \\
 \hline
\end{tabular}
\end{table}

Regarding CIFAR-10, we use a constant channel number over all the sub-blocks, exploring 3 models variants. They have 1 Yin Yang layer with 3 MBConvs and 1 single branch layer with 2 MBConvs. We apply an extra stride 2 at the first MBConv of the single branch layer on CIFAR-10 networks. The linear layer before the classification layer on CIFAR-10 has 40 neurons.

Regarding ImagetNet, we start with 16 channels, then a constant number of 2 channels is added at each MBConv on the Yin and Yang branches. After that, a fixed number of channels 64 is provided for the first single branch sub-block. We then continue adding channels after each block. We also use a single layer and 3 MBConvs for the Yin Yang branches and 2 MBConvs for the single branch sub-blocks, but in this dataset we use 4 layers of single-branch and no extra stride 2 is applied. The linear layer before the classification layer has 500 neurons.

\subsection{Results}

We conduct experiments with the small version of YYNet on CIFAR-10, in which we reach State-of-the-Art (SOTA) at model efficiency for models with few parameters. These results are provided in Table \ref{tab:cifar10}. 

\begin{table}[htb]
\centering
\caption{CIFAR-10}
\label{tab:cifar10}
\begin{tabular}{l r r} 
 \hline
 \textbf{Model} & \textbf{Test Accuracy} & \textbf{Parameters} \\ [0.5ex] 
 \hline
 ExquisiteNetV2 \cite{exquisitenetv2} &  92.52 & 890,000  \\ 

YYNet Small 64 channels (ours) & \textbf{93.32} & 726,274 \\
 
 kMobileNet 16ch \cite{kmobilenet} & 89.81 & 240,000 \\ 
 
 YYNet Small 32 channels & 91.91 & 191,330 \\
 
 YYNet Small 16 channels & 89.46 & \textbf{52,882} \\
 \hline
\end{tabular}
\end{table}

We also test a model version on the ImageNet validation dataset. We do not provide results on the test dataset, since ImageNet team only send results on the test set when the challenge is open. Comparison with similar size models are provided in Table \ref{tab:imagenet}. Our model uses 6\% less parameters than \cite{seaformer} reaching an validation accuracy only 1.2\% smaller. Regarding with MobileNet V3 \cite{mobilenetv3}, the authors do not provide validation accuracy but we assume the validation accuracy is similar to test accuracy. In this case, our model is considerable smaller and presents similar efficiency.

\begin{table}[htb]
\centering
\caption{ImageNet}
\label{tab:imagenet}
\begin{tabular}{l r r r} 
 \hline
 \textbf{Model} & \textbf{Validation Acc} & \textbf{Test Acc} & \textbf{Parameters} \\ [0.5ex] 
 \hline
 MobileNet V3 (Small) \cite{mobilenetv3} & - & 67.4
 & 2.5M  \\ 
 SeaFormer (Tiny) \cite{seaformer} & 67.7 & 67.9 & 1.7M \\
 YYNet (ours) & 66.49 & - & 1.6M \\
 \hline
\end{tabular}
\end{table}

\section{Conclusion and Discussion}

In this work, we took inspiration in neuroscience research to model efficient neural networks. We developed a two branch stem for CNNs intended to analyze colors and shapes of images separately. Our model reached State-of-the-Art results on CIFAR-10 considering models with few parameters. We reached 93.32\% test accuracy with 726k parameters, 0.8\% more than the older SOTA in this category, while having close to 150k less parameters. Our model with 52k parameters, 17 times smaller than ExquisiteNet, lose only 3.86\% test accuracy. We reached 66.49\% validation accuracy on ImageNet with 1.6M parameters.

Future work includes parameter search for the Yin Yang network at the ImageNet dataset. Yin branch would also benefits from this parameter search, as it applies stride 2 relatively late in this branch, increasing processing cost.

We also plan to investigate if our architecture is useful for other tasks beyond classification. One example is applying YYNets in generative AIs such as Stable Diffusion, by changing the latent space currently generated by U-Nets \cite{stablediffusion}. Another possible use is combining YYNets shape and color separation with architectures such as ViT, i.e, adding gray-scale input patches or queries.

\bibliographystyle{abbrv} % ver abaixo para opções
\bibliography{references} % incluir referencias.bib

\end{document}